# Thermal-Infrared Remote Target Detection System for Maritime Rescue based on Data Augmentation with 3D Synthetic Data

Sungjin Cheong, Wonho Jung, Yoon Seop Lim, Yong-Hwa Park

Department of Mechanical Engineering, Korea Advanced Institute of Science and Technology, Daejon 34141, Korea

*Abstract*— **This paper proposes a thermal-infrared (TIR) remote target detection system for maritime rescue using deep learning and data augmentation. We established a self-collected TIR dataset consisting of multiple scenes imitating human rescue situations using a TIR camera (*FLIR*). Additionally, to address dataset scarcity and improve model robustness, a synthetic dataset from a 3D game (*ARMA3*) to augment the data is further collected. However, a significant domain gap exists between synthetic TIR and real TIR images. Hence, a proper domain adaptation algorithm is essential to overcome the gap. Therefore, we suggest a domain adaptation algorithm in a target-background separated manner from 3D game-to-real, based on a generative model, to address this issue. Furthermore, a segmentation network with fixed-weight kernels at the head is proposed to improve the signal-to-noise ratio (SNR) and provide weak attention, as remote TIR targets inherently suffer from unclear boundaries. Experiment results reveal that the network trained on augmented data consisting of translated synthetic and real TIR data outperforms that trained on only real TIR data by a large margin. Furthermore, the proposed segmentation model surpasses the performance of state-of-the-art segmentation methods.**

*Index Terms*—**Infrared (IR) imaging, Remote target detection, Deep learning, Maritime rescue, Data augmentation, Synthetic data, Domain adaptation, Segmentation**

## I. INTRODUCTION

Infrared (IR) imaging has been widely used for various applications such as the military, autonomous driving, robotics [1], surveillance, and rescue. Compared to conventional RGB images, IR imaging has advantages in object detection tasks owing to its illumination and environmental robustness [2]. Especially when it comes to heat source detection tasks, thermal-infrared (TIR) images surpass the performance of RGB images [2]. Especially in maritime rescue situations, TIR imaging technology can be utilized to detect missing or drowning people. In that perspective, golden time is of utmost importance. Furthermore, with the rapid development of artificial intelligence (AI) and robots, unmanned search and rescue technology has been spotlighted. Therefore, research for a proper algorithm for maritime rescue is necessary to fulfill this objective. However, IR images inherently suffer from low contrast and unclear textile features [3], which contribute to difficulties in detecting and localizing remote targets. Additionally, IR image datasets containing high-resolution remote targets are scarce in general [4], which makes the training-based automation of object detection very difficult.

Recently, there have been a number of attempts to utilize synthetic data for real-world task solutions due to its availability [5, 6, 32]. There are several advantages of using synthetic data. First, unlike other tasks such as recognition, classification, and detection, segmentation tasks require tremendous human effort to conduct pixel-wise labeling. On the other hand, 3D games such as GTA 5 [24] support the auto-labeling function for scene segmentation. Moreover, TIR datasets with high-resolution images are scarce compared to those with visible images due to their high cost. However, by using synthetic TIR data, unlimited source data can be added to the dataset. Lastly, in terms of the maritime environment, imitating a rescue situation has some hazardous factors for a person, such as waves, winds, and water temperature. Thus, there follow a number of restrictions on acquiring datasets with various situations. Conversely, using synthetic data can eliminate those safety-related problems mentioned above. Despite these advantages, the domain gap between the synthetic images (source domain) and the real images (target domain) is a major obstacle to training-based algorithms [5, 6, 32]. Especially, unlike visible images which are capture with reflected light, the modern physics engine is somewhat inadequate to entirely reflect the principle of how thermal-infrared cameras sense the radiation in the real world (Fig. 1). Therefore, a domain adaptation algorithm is essential to reduce the domain difference.

Recent works on utilizing synthetic data for semantic segmentation propose a two-stage algorithm [5, 6, 32]: (1) a domain adaptation module and (2) a segmentation module. Specifically, the domain adaption network is trained in an unsupervised way to learn translation from the source domain to the target domain. Then, with the translated source domain, segmentation is conducted. Li *et al.* introduces a bidirectional learning framework [32] to overcome performance deterioration caused by the former step. Owing to its bidirectional learning, translation and segmentation steps promote each other alternatively in a closed loop. However, these methods only focus on generating real-like images from the source domain, ignoring the shapes and locations of small objects such as pedestrians, stop signs, and traffic lights. Regarding the principle of generative models, learning real data distributions is more significant than preserving detailed features.

In this paper, a thermal-infrared remote target detection system for maritime rescue is proposed with data augmentation following a segmentation network based on fixed-weight kernels [30]. Specifically, targets are detached from the background using ground-truth labels. Then, only the background image is flowed into the domain adaptation module in order to transfer the styles. Next, the targets are concatenated to the domain-adapted background, considering the fact that segmentation networks classify a pixel with local and global context, not only a sole pixel intensity. Lastly, the segmentation network is trained on augmented data consisting of translated TIR data (*source*) and real TIR data (*target*). The fixed-weight kernels at the head of the network enhance the image by suppressing uniform and high-frequency clutter. And at the tail of the network, a target likelihood map is generated, indicating a pixel-wise probability of the target's existence. With the augmented data, performance has improved compared to the network trained only with synthetic TIR data or real TIR data. Furthermore, our segmentation model exceeds the performance of other deep learning-based networks in terms of mean Intersection Over Union (*mIOU*) and recall with the aid of fixed-weight kernels.

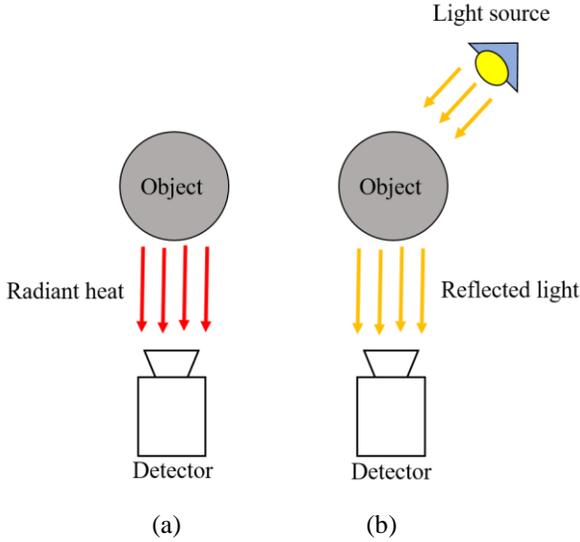

Fig. 1. Illustration of the different principles between (a) thermal-infrared, and (b) visible camera.

In summary, our key contributions are:

1. We establish a self-collected TIR dataset containing thermal-infrared images of remote human imitating maritime rescue.

2. We introduce a novel domain adaptation algorithm in order to augment the real TIR data with synthetic TIR data. By separating the targets and the background, it can even preserve the shapes and locations of the small sized targets, unlike previous methods.

3. We present a segmentation network with fixed-weight kernels at the head to suppress uniform, and high frequency noise, leading to enhancement of the signal-to-noise ratio (SNR).

## II. RELATED WORKS

### A. Segmentation network

Convolutional neural networks (CNNs) are now widely used in various tasks. From LeNet to VGG [8], [9], and the residual network [7], there has been an enormous improvement in the object detection tasks. To detect and localize the targets simultaneously, R-CNN-based methods [15-17] were proposed. However, these methods were not adequate to detect small targets due to the trade-off between the depth of the networks and the size of the receptive fields. SSD [11] was proposed to address these problems. They concatenate the features from the different layers to capture multi-scale targets. There are many variants of this multi-scale feature fusion method. Segmentation tasks require more complicated steps than recognition and detection tasks. Unlike these tasks, segmentation networks further need to classify per-pixel labels. A fully convolutional network [12] (FCN) was proposed to replace the fully connected layers in a classification task since it flattens the features, leading to the loss of spatial information. To transfer and preserve the features from the deep layers to the shallow ones, the U-Net [13] architecture with skip connections was introduced to compensate for the drawbacks of FCN [12]. The pyramidal structure forces restoration of the original dimensions. Atrous convolutions were further introduced in *DeepLab* [14] to several issues resulting from deep architecture, such as blurring, etc. Furthermore, it eliminates the problem of losing detailed information through pooling.

### B. Domain Adaptation (DA)

Image style transfer, one kind of domain adaptation, has been widely used in various applications such as image generation [18], image editing [19], and representation learning [20]. The Generative Adversarial Network (GAN) was released first in [21]. Unlike the previous encoder-decoder structure, to fully reconstruct the dimension of the original image, adversarial loss was added to fool the discriminator other than reconstruction error. Through this process, the generator learns to generate more real-like images to fool the discriminator, while the discriminator learns to discriminate real versus fake images, called the minimax problem. Pix2pix [22] is the first conditional generative adversarial network that was devised to transfer style between two images by learning a mapping function from the source to the target domain. Other than Euclidean loss, further adversarial loss is added to the transfer style as if it were real. However, in order to train the Pix2Pix network, it is necessary to have pair images, which are difficult to get. To address this problem, unpaired image-to-image translation, CycleGAN [23], was presented. The purpose of it is to transfer the source domain to the target domain and vice versa. Mathematically, a generator and another generator are trained simultaneously in a bijection way. Additionally, cycle consistency loss terms are added to encourage and not distort the structure and shape of an image, just transfer the style. Combining this loss with adversarial losses on domains yields the full objective for unpaired image-to-image translation.

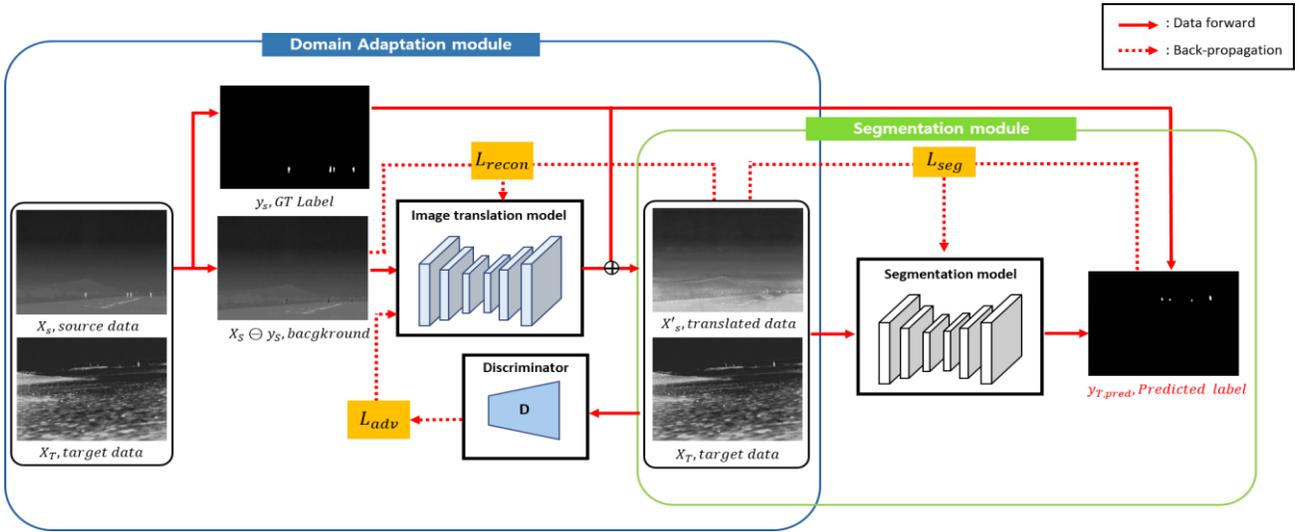

Fig. 2. The overall pipeline of the proposed algorithm. It consists of domain adaptation module, and segmentation module.

### C. Unsupervised Domain Adaptation (UDA) for segmentation

Recently, various domain adaptations have been introduced for semantic segmentation networks to overcome the laborious work of pixel-wise labeling and its enormous cost. One feasible and effective action to alleviate these efforts is to train networks on synthetic data that can be labeled automatically. For instance, GTA 5 [24] and SYHTHIA [31] are the two most popular synthetic datasets under unban environments for autonomous driving research. In this realm, domain adaptation techniques can be used to learn mapping from synthetic to real datasets. The first work for domain adaptation in the semantic segmentation task is [25], which aligns the global and local alignments between two domains at the feature level. Also, curriculum domain adaptation [26] was introduced, which estimates the distribution and the labels for the super-pixel, and then a segmentation network is trained for the finer pixel. In [27], foreground and background classes are separated and then treated for deceasing the domain gap individually.

There has been huge progress in unpaired image-to-image translation, such as in CycleGAN [23], UNIT [28], and MUNIT [29]. It can alleviate the domain difference before conducting semantic segmentation. However, one critical problem resulting from these prior works is that the performance of the work is thoroughly dependent on the quality of image-to-image translation. Once image translation fails, errors are cumulated in the following step, resulting in poor performance. To address this problem, a bidirectional learning framework [32], where both image translation and segmentation models promote each other are trained alternatively in a closed loop, is introduced. Specifically, segmentation and domain adaptation networks are trained in an alternative way to promote each other. Nonetheless, even if it can perfectly segment large structures such as cars, buses, and small objects like pedestrians, stop signs are often ignored since generative adversarial networks are only aimed at generating real-like data. Therefore, in this paper, the target-background-separated domain adaptation technique for semantic segmentation is used to address these problems. We first separate the small targets from the background, apply unsupervised domain adaptation to the background, and then attach the targets to the background to preserve the small objects.

## III. PROPOSED REMOTE TARGET DETECTION ALGORITHM

In this paper, a two stage algorithm (Fig. 2), unsupervised domain adaptation module followed by a segmentation module [30], is presented. In the domain adaptation module, the domain of 3D game TIR data is transferred into that of real TIR data with target-background separation to preserve the small size of the objects. In order to augment the data, synthetic data is utilized (Fig. 3). Additionally, the proposed segmentation module consists of three parts [30]. First, fixed-weighted kernels are utilized for enhancing the targets and suppressing the noise beforehand, given that the background and the clutter are usually uniform. Then, a U-Net like structure with an encoder-decoder based on *Resblock* [7] reconstructs the input image. Skip connections were used to transfer the features from the shallow layers to the deep layers in order to prevent these features of small objects from vanishing as it went deeper. Atrous Spatial Pyramid Pooling (ASPP) [10] is inserted between the encoder and the decoder to capture the global context. Lastly, the convolution block maps the reconstructed image to the probability map, which indicates per-pixel likelihood of whether this pixel belongs to the target or not.

### A. Maritime Rescue TIR Data Acquisition

To establish maritime rescuer data, a M364C (*FLIR*) thermal-infrared camera that uses mid-wavelength IR (MWIR, 3~5 $\mu m$ range) and long-wavelength IR (LWIR, 7.5~14 $\mu m$ range) was used. Officially, it can capture people in the water up to 823 m and small vessels up to 2,200 m. We imitate two situations of maritime recue: rescuers in the water and rescuers

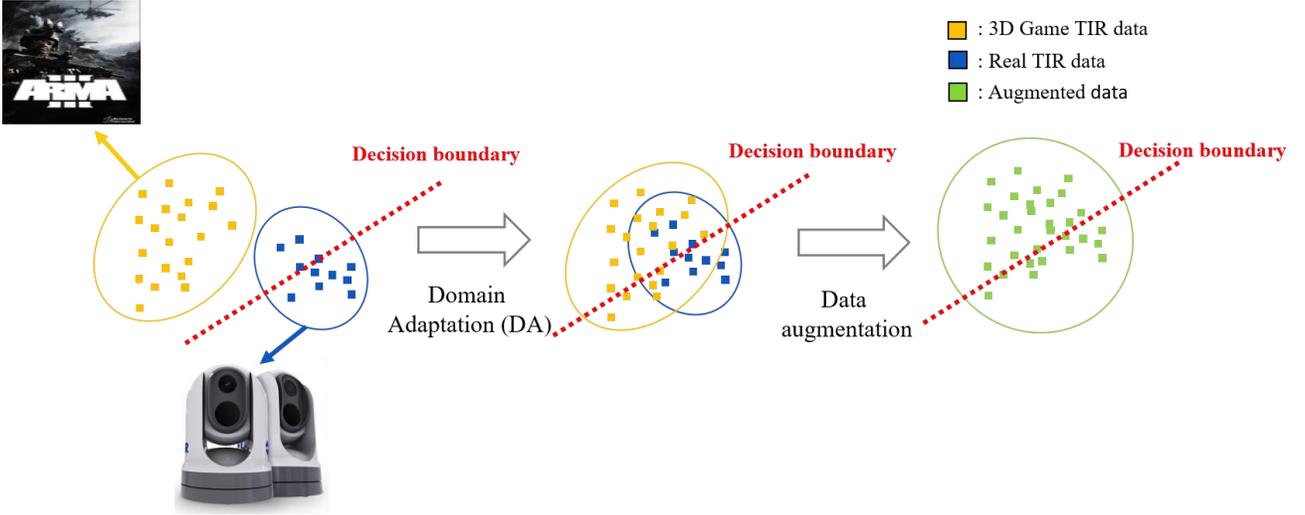

Fig. 3. The schematic view of data augmentation process. The domain of synthetic data (3D game) is transferred to that of real data.

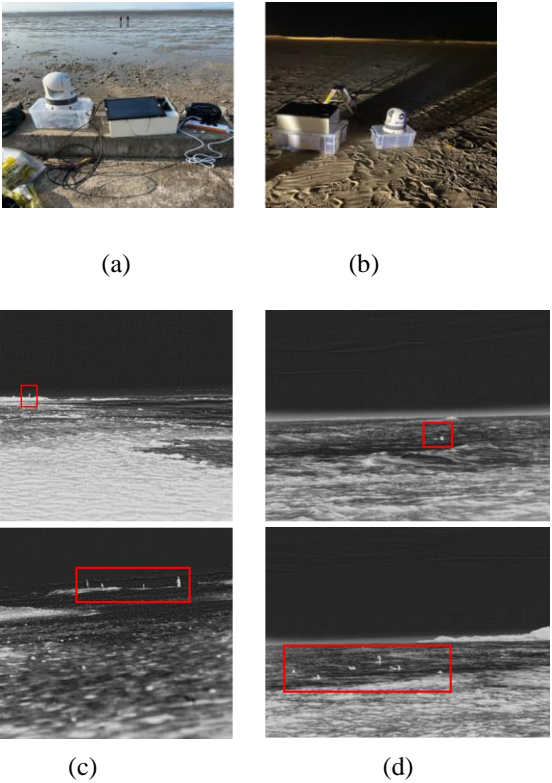

(a)                          (b)

(c)                          (d)

Fig. 4. Illustration of the self-collected dataset: Data acquisition setup at (a) day time, and (b) night time. Sample data capture on (c) the mudflat, and (d) the ocean (Red boxes indicate the targets)

on the mudflat. 8,172 images were collected for the first situation and 9,317 images for the second situation. Conclusively, total of 17,489 images were collected with 150 independent scenes with 4 different locations. The distance was measured in the range of 50 m to 500 m. Fig. 4 shows the experiment setup for data acquisition and some samples of collected data. To fully maintain the independence of intra-scene and inter-scene, various times and locations were chosen.

### B. Synthetic TIR Data Acquisition

To tackle the problem resulting from the data scarcity, further 3D games, ARMA 3 (*Bohemia interactive*), and TIR data were collected (Fig. 5). It consists of a total of 2,093 images (21 independent scenes), of which 315 images were labeled through the open-source tool, *Label-Studio* [26]. In order to imitate maritime situations, various game backgrounds were chosen, such as rivers, lakes, oceans and the beaches.

### C. The Proposed Domain Adaptation Module

To address the problem due to the data scarcity, in a further 3D game, ARMA 3 (Bohemia interactive), TIR data was collected for the purpose of data augmentation. First, an image translation model is trained with 2,093 images of 3D game TIR data (*source*) and the same number of real TIR data (*target*) as input (Fig. 2). CycleGAN [23] was chosen as a backbone network for image translation since it shows remarkable performance when it comes to style transfer in an unsupervised way. As Fig. 6 shown, while training, two generators, $P : \mathcal{S} \rightarrow \mathcal{T}, Q : \mathcal{T} \rightarrow \mathcal{S}$, and corresponding two discriminators, $D_S$ and $D_T$, are updated by the minimax algorithm. $D_S$ is the discriminator from the source domain $\mathcal{S}$, and $D_T$ is from the target domain $\mathcal{T}$. With the adversarial loss from each domain, generator $P$ and $Q$ translate each domain to the other that is indistinguishable from the translated domain. The adversarial losses are given as:

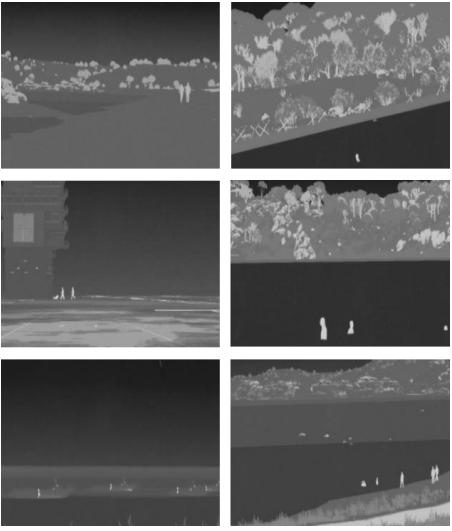

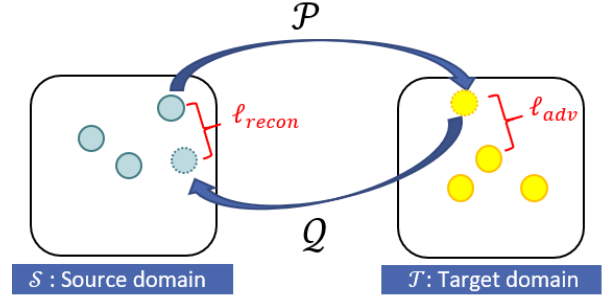

Fig. 6. A schematic view of domain adaptation process from the source to the target, and vice versa. The generator and discriminator are trained with adversarial, and reconstruction error ($\mathcal{P}$ denotes generator from source to target domain, $\mathcal{Q}$ indicates inverse function)

Fig. 5. Samples of the collected synthetic TIR data from the 3D game. It contains various backgrounds such as mudflat, river, ocean, and aircraft carrier etc.

$$\ell_{adv}(P(\mathcal{S}), \mathcal{T}) = \mathbb{E}_{X_T \sim \mathcal{T}}[D_S(X_T)] + \mathbb{E}_{X_S \sim \mathcal{S}}[1 - D_T(P(X_S))], \quad (1)$$

$$\ell_{adv}(\mathcal{S}, Q(\mathcal{T})) = \mathbb{E}_{X_S \sim \mathcal{S}}[D_T(X_S)] + \mathbb{E}_{X_T \sim \mathcal{T}}[1 - D_S(Q(X_T))]. \quad (2)$$

Where $X_S$ denotes images from the source domain, $X_T$ from the target domain.

Moreover, the reconstruction losses are added to regulate the generators to translate images, thereby at least preserving the original image. Specifically, translated images from the other domain are forced to be able to be returned to the original domain by inverse mapping [23]. In that way, the image translation model can only learn style mapping, not transform into totally new images. The reconstruction losses are defined as follows:

$$\ell_{recon}(Q(P(\mathcal{S})), \mathcal{T}) = \mathbb{E}_{X_S \sim \mathcal{S}}[\left\| Q(P(X_S)) - X_S \right\|_1], \quad (3)$$

$$\ell_{recon}(\mathcal{S}, P(Q(\mathcal{T}))) = \mathbb{E}_{X_T \sim \mathcal{T}}[\left\| P(Q(X_T)) - X_T \right\|_1]. \quad (4)$$

The total loss, $L_{trans}$, is given as the summation of the two adversarial losses and the two reconstruction losses. With this objective function, an image translation model can be trained to transfer the style of real TIR data to that of 3D game TIR data.

$$L_{trans} = \ell_{adv}(P(\mathcal{S}), \mathcal{T}) + \ell_{adv}(\mathcal{S}, Q(\mathcal{T})) + \ell_{recon}(Q(P(\mathcal{S})), \mathcal{T}) + \ell_{recon}(\mathcal{S}, P(Q(\mathcal{T}))). \quad (5)$$

After training, target-background separated domain adaptation is applied. Since a generative model learns data distribution, objects of small size are often distorted, transformed, and eliminated. Unlike prior works, which focused on only generating real-like synthetic images, this paper presents a novel domain adaptation technique in a target-background separated manner. As Fig. 2 shown, separated background is flowed into the trained image translation model. After that, the target masks are added, resulting in translated data, $X'_S$. The detailed procedure of target-background separations is as follows: First, if a pixel belongs to a target label, all pixel intensities in a row containing that pixel are calculated. Then, the average value of the row is filled into a pixel. In this way, all of the target pixel values are replaced with averaged row pixel values. By separating the target and the background, the shape and location of the targets can be preserved. Even if only background is translated into the target domain, it does not affect the total performance since segmentation models classify per-pixel labels not by sole pixel value, but by the local and global context of a scene.

### D. The Proposed Segmentation Module

The next step is the segmentation module (Fig. 7). A novel segmentation aimed at infrared small object detection is presented in our previous work [30]. Overall architecture is as shown in Fig. 7. It consists of three modules, sequentially: image enhancement, multi-scale feature extraction, and probability estimation. Through the image enhancement module, fixed-weighted kernels are convoluted with the input image with different aspect ratios to improve SNR and give weak attention at the head. Next, U-Net structure extracts features and reconstructs the original image dimension. The Atrous Spatial Pyramid Pooling (ASPP) module is added between the encoder and the decoder to further capture contexts. Lastly, the convolutional block is convoluted with the output of the decoder to map per-pixel target likelihood. The real TIR data augmented with translated source domain image is flowed into the segmentation model, which is then trained with binary cross-entropy loss to estimate target likelihood.

Since a target is far away and immersed in heavy clutter, its signal-to-noise ratio (SNR) and local contrast are extremely low, which makes it difficult for a deep neural network to extract features from the beginning. Thus, fixed-weighted kernels are inserted at the head to enhance the target and suppress the uniform noise in advance to help the deeper automated kernel to learn patterns for a small target easily. Furthermore, kernels with different aspect ratios were used to capture the unpredicted shapes of a target without losing generality. The Fig. 8 shows the fixed-weight kernels with different aspect ratios and the schematic view of how there are convoluted with the input image. Specifically, the red cells, denoted as $K_r$, and the blue cells, $K_b$, are convoluted with the original image, $I(x, y)$, where $x$ and $y$ indicate the coordinates of the pixel in an image, are convoluted as follows:

$$I'(x', y') = \sum_R K_r(x, y) \circ I(x, y) - \sum_B K_b(x, y) \circ I(x, y), \quad (6)$$

$$K_r(x, y) = \frac{1}{n_r}, \ K_b(x, y) = \frac{1}{n_b}. \quad (7)$$

where $n_r$ is the number of red cells in the kernel and $n_b$ is that of blue cells. $I'(x', y')$ is the output of the enhanced image after the convolutional operation, defined as $\circ$. The total value of red cells is subtracted from that of blue cells, whose summed values equal one. Ablation studies about a performance discrepancy with fixed-weighted kernels and with free-weighted kernels (normal convolutional kernels) are conducted in later section.

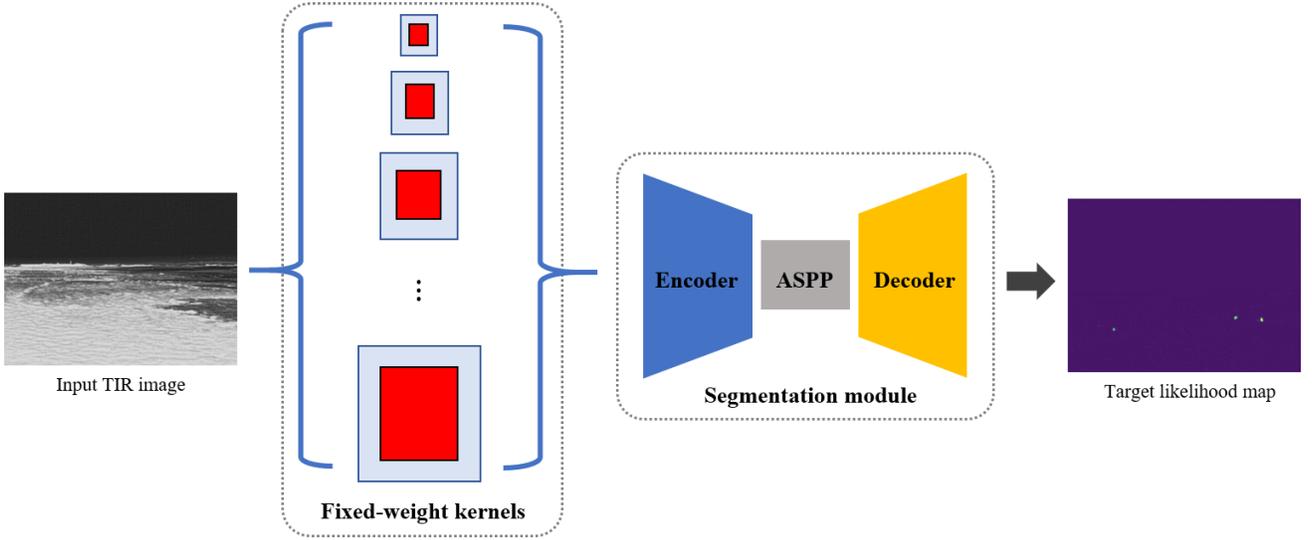

Fig. 7. The flowchart of the utilized segmentation algorithm. Input TIR image is convoluted with fixed-weight kernels to enhance the image, then flowed into the encoder-decoder structure, resulting to target likelihood map [30].

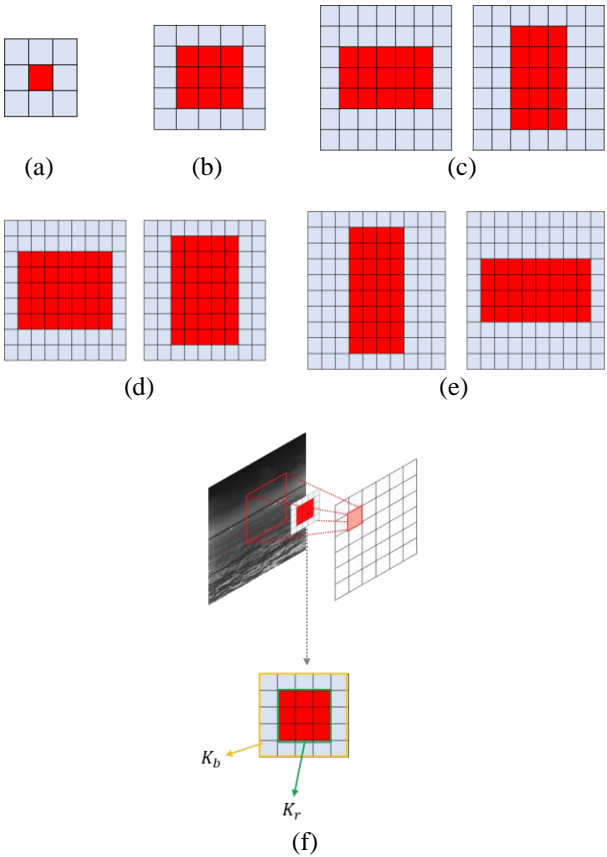

Fig. 8. Detailed illustrations of the fixed-weight kernels with different kernel size and aspect ratio. Kernel size with (a) 3×3, (b) 5×5, (c) 7×7, (d) 9×9, and (e) 11×11. (f) shows the process of convolution, where $K_b$ denotes blue cells while $K_r$ indicates red cells

A U-Net like encoder decoder structure with residual blocks is used as a backbone network to reconstruct the input image. Residual blocks transfer the features from layer to layer, preventing the features from vanishing and helping gradient flow by backpropagation. Additionally, skip connections between the shallow and deep layers can further block the loss of features resulting from a large receptive field in the deep layers. Between the encoder and the decoder, the ASPP module is inserted to capture the global context of a scene and help to grasp the targets at different scales. ASPP was first introduced in *Deeplab v3+* [10] and, by far, has been used in a large

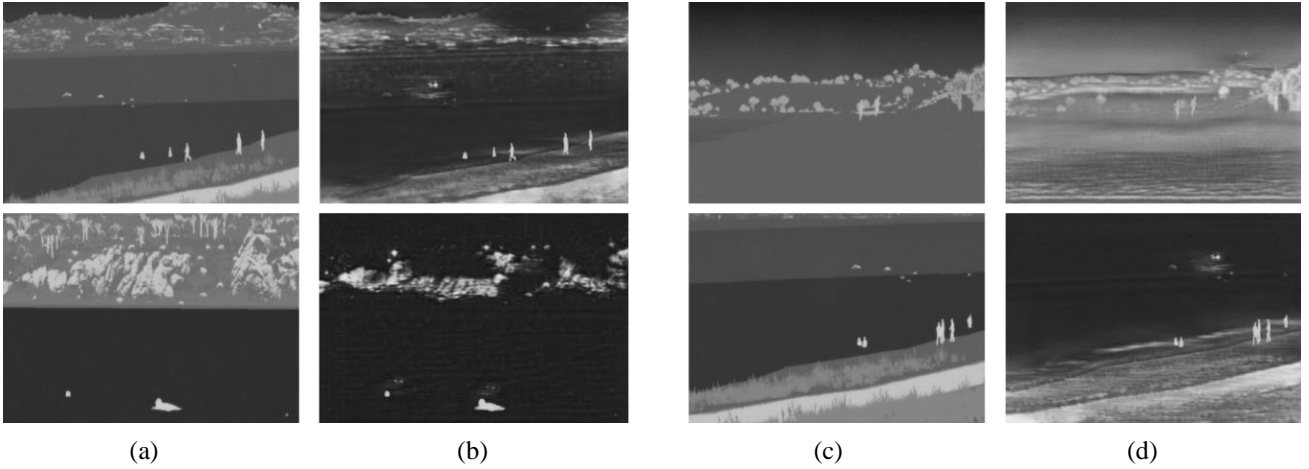

| (a) | (b) | (c) | (d) |

Fig. 9. The results of domain adaptation from the 3D game TIR images to the real TIR images with target-background separation. (a), (c) denote 3D game TIR images, and (b), (d) indicate translated TIR images

number of segmentation tasks where multi-scale context information is required. In ASPP, atrous convolution is utilized, which has the advantage of requiring smaller parameters compared to normal convolution with the same receptive field. Furthermore, it can capture different scales of objects with various dilation rates, preserving spatial and semantic information. In this work, four kernels with different dilation rates were used to capture local context and multi scale objects. Additionally, average pooling layers are added to capture global context and large clutter such as waves, land, or sky in a scene. Finally, the features from atrous convolutions and average pooling are concatenated and then flowed into the decoder as input.

The final output of the decoder flows into a convolution block, *Conv*, of which kernel size is 1, resulting in a target likelihood map after the *softmax* layer. The target likelihood map consists of two channels. The first and second channel indicate how likely a pixel belongs to the background or the target, respectively, with a value of 0 to 1. In other words, the network is trained to classify each pixel of the input into two categories, targets, and background. The final target probability map is generated after the binarization process, in which a pixel of which target probability is higher than the background probability is 1 or else 0. Specifically, the output of the decoder, $Z$, is flowed into the convolutional block, leading to the target likelihood, $P_t(x, y)$, and background likelihood, $P_b(x, y)$, where $x, y$ are the coordinates of the pixel in a feature map. The formula is given as follows:

$$P_t(x, y), P_b(x, y) = Softmax(Conv(Z(x, y))), \quad (8)$$

$$P_t(x, y) + P_b(x, y) = 1, \ (P_t(x, y) \geq 0, P_b(x, y) \geq 0). \quad (9)$$

Target likelihood binary map, $P$, is generated given as:

$$P(x, y) = \begin{cases} 1, & P_b < P_t \\ 0, & P_b \geq P_t \end{cases}. \quad (10)$$

### E. Training Details

For domain adaptation model training, 2,093 images of 3D game TIR data and the same number of real TIR data were trained in an unsupervised way with 200 epochs. Two adversarial losses and two reconstruction losses were used as an objective function to update two generators and two discriminators. For the generators, the U-Net structure was utilized as a backbone and the 70x70 *PatchGAN* [22] structure as discriminators. In the case of segmentation module training, a total of 275 labeled images from real TIR data, augmented with 315 images of domain adapted 3D game TIR images, were utilized. For real TIR data, 75 images were divided as a test and the rest as a train. To fully validate the effect of data augmentation, 3D game TIR data was not included in the test data. Moreover, since the total amount of dataset is not enough to calculate robust performance, k-fold cross validation was applied to real TIR data. Average values of *mIOU*, recall, precision, and f1 score were calculated from 4-folds. The experiment was conducted on an environment of 20.04, *Ubuntu*, *RTX 3090 Ti*.

## IV. RESULT AND ANALYSIS

### A. The Target-Background Separated Domain Adaptation

Fig. 9 above shows the source data (3D game TIR images) and the translated source data using the proposed domain adaptation algorithm. The thermal-infrared image sensor is vulnerable to humidity, temperature, and the various camera parameters, resulting in a low signal-to-noise ratio that generates uniform noise across the image. On the other hand, 3D game TIR data is a virtual image that is not affected by these factors, thereby giving a clear view of the image. However, by

## TABLE I

COMPARISON OF MIOU, RECALL, PRECISION, AND F1
SCORE BETWEEN THE PREVIOUS DEEP LEARNING-BASED
SEGMENTATION NETWORKS AND THE PROPOSED METHOD

| | mIOU (%) | Recall (%) | Precision (%) | F1 Score (%) |
|---|---|---|---|---|
| DA + FCN | 89.4 | 84.6 | 92.5 | 88.4 |
| DA + PSPNet | 91.9 | 87.6 | 95.3 | **93.1** |
| DA + U-Net | 92.0 | 90.9 | 92.1 | 91.5 |
| DA + SegNet | 89.8 | 85.5 | **92.7** | 88.9 |
| DA + DeepLab v3 (Backbone: MobileNet) | 82.6 | 84.6 | 72.9 | 78.3 |
| DA + DeepLab v3 (Backbone: Xception) | 74.8 | 62.4 | 73.3 | 67.4 |
| **DA + Proposed** | **93.6** | **97.8** | 88.9 | **93.1** |

## TABLE II

COMPARISON OF MIOU, RECALL, PRECISION, AND F1
SCORE BETWEEN THE NETWORK TRAINED WITH ONLY
TRANSLATED 3D GAME TIR DATA, ONLY TIR DATA, AND
BOTH

| | mIOU (%) | Recall (%) | Precision (%) | F1 Score (%) |
|---|---|---|---|---|
| 3D Game TIR | 82.2 | 88.2 | 73.2 | 80.0 |
| Real TIR | 92.7 | 92.1 | 92.2 | 92.1 |
| 3D Game TIR + Real TIR | **93.1** | **92.3** | **93.1** | **92.7** |

## TABLE III

COMPARISON RESULTS OF THE NETWORK TRAINED
WITH SOURCE DATA (3D GAME TIR DATA) AND
TRANSLATED SOURCE DATA

| | mIOU (%) | Recall (%) | Precision (%) | F1 Score (%) |
|---|---|---|---|---|
| Trained with source data | 82.6 | 79.6 | 76.7 | 78.1 |
| Trained with Translated source data | **86.7** | **83.5** | **85.4** | **84.4** |

domain adaptation process, 3D game TIR domain is translated into real TIR domain which generates real-like noise and clutter, eliminating artificial texture. Moreover, domain adaptation in a target-separated manner gives remarkable performance without losing or distorting the shapes of small targets.

### B. Comparison with Other Deep Learning-based Segmentation Network

To validate the performance of the proposed segmentation network, other deep learning-based segmentation networks are compared. Sequentially, Fully Convolutional Network (FCN), PSPNet, SegNet, U-Net, and DeepLab v3+ [12,27,33,13,10] with the backbone network of Mobilenet [34], and Xception [35] are presented, respectively. In the first scene (Fig. 10), due to the small size of the target, false positive (FP) and false negative (FN) samples are generated. However, the proposed method can perfectly estimate the target probability. As Fig. 11 shown, under the multi-target situation, even the other deep learning methods cannot detect small-sized targets in the scene, the proposed method can even segment the multi-targets including small-sized objects, from the background. Additionally, when there is a huge wave of which pixel intensity is similar to that of humans (Fig. 12), the proposed method can estimate target probability and localize the target even under heavy clutter. Four different qualitative metrics were used to measure the performance of detectability. First, mean Intersection Over Union measures the localization performance of a network. The metric is defined as:

$$mIOU = \frac{1}{N}\sum_{i}^{N} IOU_i, \quad (IOU_i = \frac{Intersection}{Union} \text{ of } i^{th} class). \quad (11)$$

On top of that, recall, and precision further were utilized, given as:

$$Recall = \frac{TP}{TP+FN}, \quad (12)$$

$$Precision = \frac{TP}{TP+FP}. \quad (13)$$

Lastly, the F1 score metric is added since recall and precision are trade-off relations. Specifically, depending on the extent of the threshold determining whether a pixel is classified as a target or a background, the values of recall and precision vary in an inverse way. Therefore, the harmonic mean of the two values is used to have a reliable measurement regardless of threshold. The f1 score is defined as:

$$F1 \ score = \frac{2 \cdot Precision \cdot Recall}{Precision + Recall}. \quad (14)$$

As shown in Table I, the proposed segmentation network surpasses the other deep learning-based methods in terms of *mIOU* and recall, rising by 1.7% and 6.9%. respectively. As the fixed-weight kernels work as both weak-attention modules and image enhancement at the network head, recall has increased dramatically [30].

### C. Data Augmentation Using Synthetic TIR Image Data

To demonstrate the effect of data augmentation, we conducted three experiments, training with real data only, 3D game data only, and both. A total of 315 images of 3D game TIR images and 275 images of real TIR images were used.

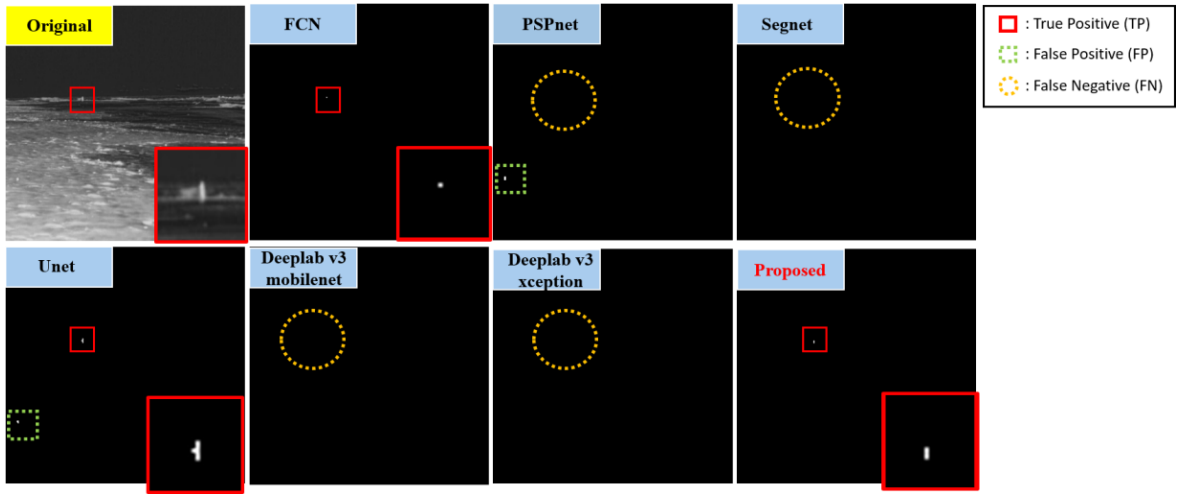

Fig. 10. The inference results of the 1ˢᵗ scene using other deep learning-based segmentation, and the proposed algorithm.

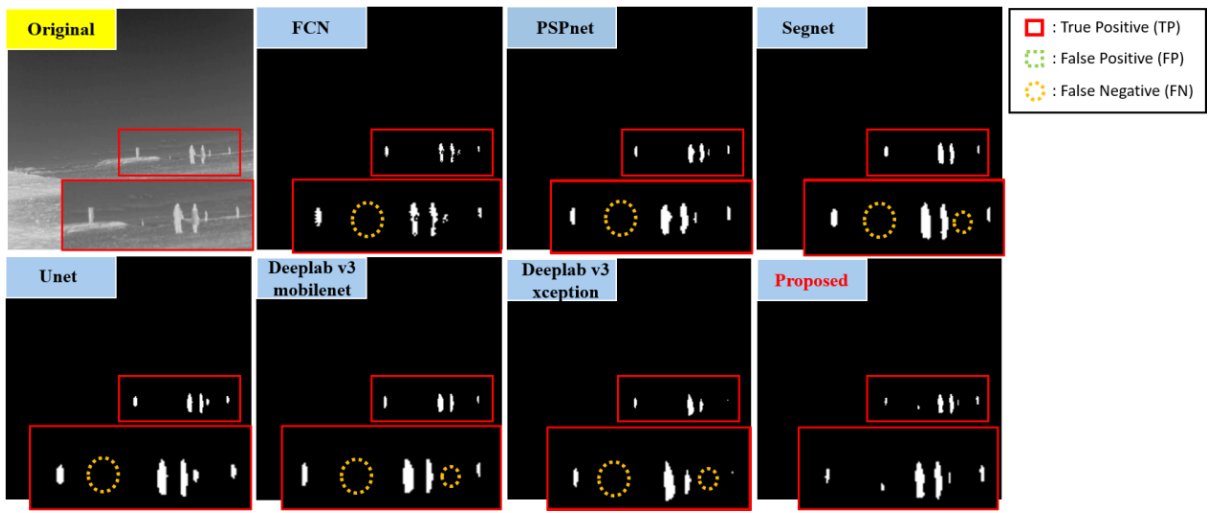

Fig. 11. The inference results of the 7ᵗʰ scene using other deep learning-based segmentation, and the proposed algorithm

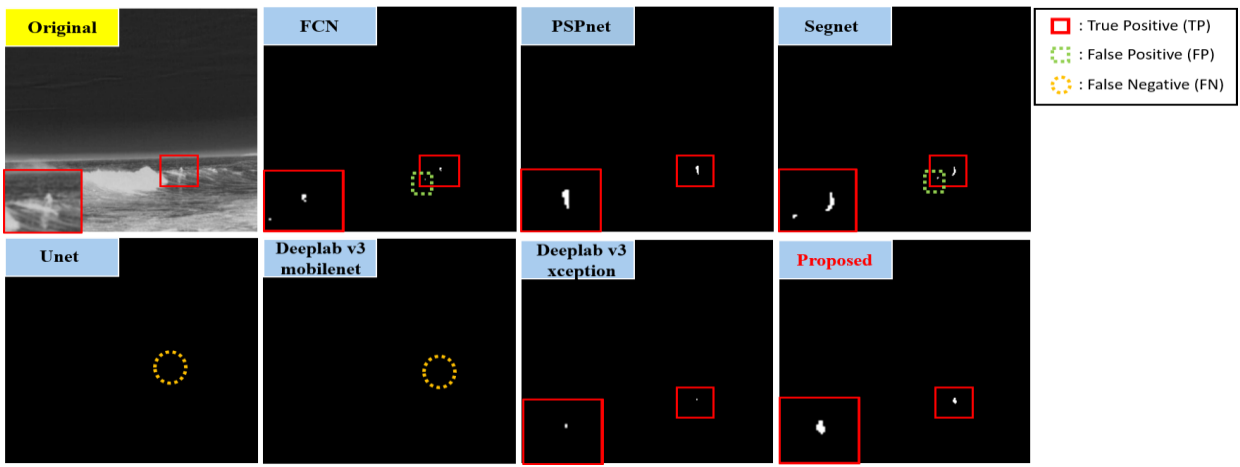

Fig. 12. The inference results of the 13ᵗʰ scene using other deep learning-based segmentation, and the proposed algorithm

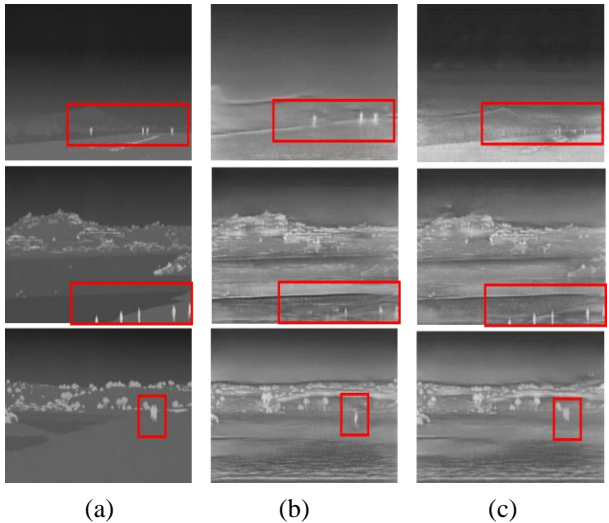

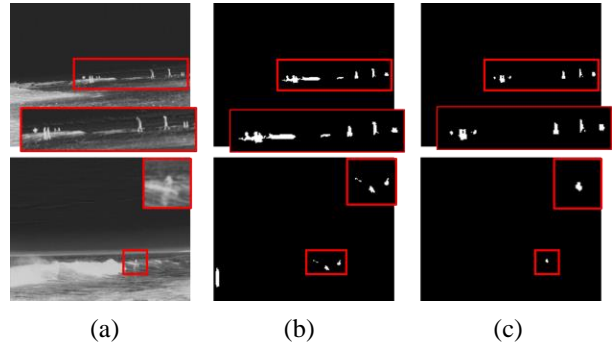

Fig. 14. The inference results of the network trained on with and without domain adaptation from the source to the target. (a) Original images, (b) output of the network trained without domain adaptation, and (c) with domain adaptation (Red boxes indicate the targets, and enlarged images of those)

Fig. 13. The results of domain adaptation with and without target-background separation. (a) Original images, domain transferred images (b) without separation, and (c) with separation (Red boxes indicate the targets)

In real TIR images, 75 images were set as a test and the other as a train. Since the number of datasets is small, k-fold cross validation was used to measure a reliable qualitative result. Specifically, real TIR images were divided into four folds, calculating the metrics for each fold as a test set. Then, the average value of the summation of each fold is measured. Table II shows the comparison results with and without data augmentation. The lowest performance was when the network was trained with only 3D game data. Even if the 3D game domain is translated into real domain, there still exits the discrepancy in data distribution. Nevertheless, the network trained with 3D game and real data together shows slightly higher performance than the network trained with only real data in terms of every metric. This indicates that without collecting a large amount of real-world data, the network trained with synthetic data can show sufficient performance.

## V. ABLATION STUDIES

### A. *The Effect of Target-Background Separated Domain Adaptation*

As Fig. 13 shown, domain adaptation without target background separation shows poor performance. The shape and location of the small objects are distorted, transformed, and even removed. Since the generative model learns the data distribution, it only focuses on generating real-like images with relatively huge structures. Therefore, small objects are often ignored, leading to segmentation performance degradation. Moreover, to validate the effect of domain adaptation, an experiment where the network is trained on source domain (3D game data) with and without domain adaptation is conducted. As Fig. 14 shown, the results trained without domain adaptation show degraded performance, generating several false positive samples. From Table III, the performance of trained with translated source data exceeds that of trained with source data in terms of *mIOU*, recall, precision, and f1 score.

### B. *The Effect of Fixed-Weight Kernels*

Furthermore, fixed-weight kernels are compared with free-weight kernels of which weight values need to be trained. A comparison was conducted with the same size and number of kernels at the head of the backbone network. As Table IV shown, the network trained with fixed-weight kernels outperforms that trained with free-weight kernels in terms of the all metrics other than precision. The probable reason for decreased precision can be explained by the fact that fixed-weight kernels localize the bright and round shapes of the objects. In that sense, there appears to be a possibility of similar objects being captured by these kernels, leading to the generation of more positive samples. Nonetheless, in this work, the risk of deciding human as background is regarded as much more serious than that of deciding background as human since the former case could cause significant maritime casualties. Thus, higher recall rates with a lower precision algorithm are considered proper for this objective.

## VI. CONCLUSIONS

In this paper, a novel infrared small target detection algorithm was proposed comprising a domain adaptation module, fixed weight kernels, an ASPP module, and residual UNet. Considering the objective of maritime rescue situations, higher recall rates are much more valued than precision rates since these two values are trade-offs. Thus, raising recall rates is the main focus of this paper. A domain adaptation algorithm was applied for 3D game-to-real conversion based on target-background separation. The network trained with augmented

TABLE VI

ABLATION STUDY RESULTS OF THE NETWORK CONSISTING OF THE FIXED-WEIGHT KERNELS AND FREE KERNELS UNDER THE SAME CONDITION

| | mIOU (%) | Recall (%) | Precision (%) | F1 Score (%) |
|---|---|---|---|---|
| Fixed-weight kernel | **93.1** | **92.3** | 93.1 | **92.7** |
| Free kernel | 91.8 | 88.8 | **93.7** | 91.2 |

data from 3D game TIR data shows improved performance compared to that of real TIR data. Furthermore, with the aid of target-background separation, the shapes and locations of the small objects were maintained, transferring only style. In the proposed segmentation module, fixed-weight kernels aid the network in extracting features easily beforehand. Moreover, in the ASPP module, dilated convolution kernels with different dilation ratios can capture multiscale features, and average pooling can grasp global context. From the results, the proposed method can perfectly localize and infer the likelihood of the target's existence per pixel, regardless of the shapes, sizes, and environment. We compared our method with the other methods and also ablation studies regarding fixed-weight kernels and the ASPP module. Nevertheless, there is still much room to improve, such as inference time, distinguishing individual instances, or tracking.

In future work, time information can also be considered since each frame is not independent, thereby reducing inference time and errors dramatically. Even if it shows improved performance, a lack of labeled data contributed an unreliable performance which is difficult to represent the whole dataset. Therefore, more labeling tasks for the dataset are required. Moreover, this research focuses on only maritime rescue situations. Recently, unsupervised domain adaptation for segmentation has been popular in the realm of autonomous driving. In this domain, detecting small objects such as pedestrians, and stop signs is of utmost importance to avoid unpredictable accidents. Therefore, it is assumed that a target-background separated domain adaptation algorithm can be utilized to improve the detectability of small objects.